# A Foundry of Human Activities and Infrastructures

Robert B. Allen, Eunsang Yang, and Tatsawan Timakum

Yonsei University, Seoul, South Korea
rballen@yonsei.ac.kr, esy220@nyu.edu, tatsawan@gmail.com

**Abstract.** Direct representation knowledgebases can enhance and even provide an alternative to document-centered digital libraries. Here we consider realist semantic modeling of everyday activities and infrastructures in such knowledgebases. Because we want to integrate a wide variety of topics, a collection of ontologies (a foundry) and a range of other knowledge resources are needed. We first consider modeling the routine procedures that support human activities and technologies. Next, we examine the interactions of technologies with aspects of social organization. Then, we consider approaches and issues for developing and validating explanations of the relationships among various entities.

**Keywords:** Community Models, Digital Humanities, Direct Representation, Faceted Ontologies, Histories, Social Science, Tangible and Intangible Culture Heritage

## 1 Infrastructures, Human Activities, Community Models, and Cultures

In [1] and related studies we explored indexing digitized historical newspapers. It was difficult to index the articles for retrieval or, even, to unambiguously identify what text should be treated as an article. Thus, we proposed the development of knowledge-rich "community models" to improve retrieval. Many aspects of infrastructure associated with everyday activities and infrastructure generally can be described with such community models. Such models would cover both tangible and intangible cultural heritage such as pottery, clothing, dance, and religious traditions.

This work is parallel to a proposal we have made for direct representation of scientific research results [4]. However, there are additional challenges for descriptions of culture and history because of the lack of consensus about the definitions for social entities and because there are disagreements about the details of cultures and histories. Nonetheless, as information scientists, we believe that it is useful to develop frameworks for articulating and exploring the possibilities. Ultimately such frameworks should support tools both for the public and for scholars.

## 2 Ontologies and Models

### 2.1 Upper Ontologies and the Model Layer

The knowledge representation system for direct representation must go beyond simple linked data to incorporate structured knowledge from many domains. To provide a framework we use an upper or formal ontology, specifically, the Basic Formal Ontology (BFO) [7]. BFO is widely applied for biomedical ontologies. It is a carefully designed, realist ontology which follows Aristotle in distinguishing between Universals and Particulars. BFO also distinguishes between Continuants (those Entities which are constant across time) and Occurrents (those Entities which change across time).

We have proposed extending BFO with a Model Layer [5] that gives it aspects of object-oriented modeling.[1] The Model Layer considers Thick Entities, which are Independent Continuants, together with their associated Dependent Continuants, Parts, and Processes [2]. Such Thick Entities should allow for States and State Changes. Such States and State Changes would not be first-class ontology entities; rather, they can be defined as derived entities [5]. If they were formalized to include State Changes, Processes could be considered as analogous to "abstract methods" in object-oriented programming. The Model Layer would also include Mechanisms and Procedures [3]. The Model Layer should be able to show how Thick Entities interact with other Thick Entities much as the objects in an object-oriented program interact when the program is executed.

### 2.2 Foundries as Knowledgebases and Extended Foundries

The Open Biomedical Ontology (OBO) Foundry [22] is a large curated collection of domain ontologies and partonomies based on the BFO.[4] We propose the development

---

[1] In some cases, "object-oriented" simply means entity or object-based. We use "object-oriented" in the stronger, programming-language sense of objects that include specific processes and procedures.

[2] The descriptions of Thick Entities we envision are analogous to the descriptions of Model or Reference Organisms. The latter often includes anatomies (i.e., partonomies) and, less often, descriptions of related Procedures and Mechanisms.

[3] A Mechanism describes how a Process is implemented. A Procedure is like a workflow with flow control and decision points. There is no direct way for BFO to represent control statements such as loops and conditionals needed for Procedures, although it is possible to represent control statements with OWL on an ad hoc basis and to use those representations in combination with BFO. A pure BFO modeling language should be developed that, like the C programming language, is self-compiling.

The distinction in some object-oriented languages between "private methods" and "public methods" can also be applied to Thick Entities. Private methods are those which interact internally only with other Parts of a given Thick Entity whereas Public Methods support interaction with other Thick Entities.

[4] obofoundry.org, obofoundry.org/docs/OperationsCommittee.html

of a similar foundry to cover human activities and infrastructures. Because "direct representation" follows BFO as a realist approach, our focus is first on describing infrastructure objects, their use and their interaction with other objects. The role of entities in collections and records are important but secondary [6].

## 3 Material Technologies and Infrastructure

The contents of historical newspapers and other historical records for small towns often describe entities and activities which are routine, even mundane. A partial list of such entities and activities includes roads, farming, fishing, blacksmithing, weaving, coin minting, pottery making, and bookbinding. Each of these is associated with specific types of objects and procedures.

There are many levels for representing and modeling everyday human activities. At a general level, we might describe infrastructures and technologies for supporting basic human needs such as food and shelter. Such models could be increasingly refined as they are applied to specific scenarios. While Aristotle focused on Universals as natural entities, BFO has included human artifacts related to scientific research such as flasks. We further extend the scope of Universals to include all types of human artifacts.

As noted above, we also propose using model-oriented Thick Entities for these descriptions. Thick Entities would include Processes and Procedures. There is a complex web of interactions in Processes and Procedures. For instance, farming procedures are affected by the availability of different metals for plows. Similarly, the introduction of a train line may dramatically affect a community (cf., [1]).

The development of a large and internally consistent collection of infrastructure entities will require a major effort that is in its early stages. Ontologies and other controlled vocabularies have been developed for many entities and functions; for instance, FGDC (fgdc.gov) provides descriptions for highways. Similarly, standard descriptions for Mechanisms and Procedures such as from the *Handbook of Synthetic Organic Chemistry* could also be included. Some aspects of Human Activities and Infrastructures (such as farming or silkworm cultivation) could be linked in the OBO. Ultimately the foundries should be unified.

## 4 Structures and Activities within a Given Social Framework

While the previous section focused on material technologies and infrastructures, ultimately, it will not be possible to separate the technologies and infrastructures from their interaction with social activities. Social structures may be considered as entities in a social ontology.[5]

---

[5] Smith (Social Objects, http://ontology.buffalo.edu/socobj.htm) claims that social entities are entirely consistent with the BFO framework. There has been significant work on social ontology by some of the designers of the BFO framework but there does not yet seem to have been a concerted effort to directly integrate that work into the BFO. Much of the discussion about social

There are many examples of the interaction of material infrastructure with social entities. For example, textiles play an important role in traditional Thai society [8]; the fabrics are integral to courtship, marriage, death, and a variety of Buddhist rituals. A structured description of the materials and technologies would include aspects of fabrics and weaving tools and techniques as well as their role in society.

In addition to tangible cultural heritage such as Thai silks, some cultural heritage like dance and music can be both tangible and intangible. On one hand, musical instruments are Continuants but musical performances are Occurrents. Moreover, music also has a social dimension. For example, descriptions of Korean music (gugak) need to include social distinctions between different genres (e.g., folk music vs. court music) [16].

In many contexts, the models of the social framework would generally be from the perspective of the participants. For historical local newspapers [1], we would generally follow the presentation of the newspaper editors in developing models of schools, businesses, government, churches, and families. Of course, there are frequently alternative interpretations beyond the normative descriptions. Therefore, flexible frameworks would need to be developed to present and contrast differing viewpoints.

## 5 Explanations and Social Science

We have described how material infrastructures are interrelated and dependent on each other. Beyond those simple descriptions, we can explore claims about the relationships among components of the material and social infrastructures at the level of both Universals and Particulars. However, in many cases, the relationships are complex and not susceptible to proof. For instance, culture can be described as a web of relationships [11]. We need to develop a flexible framework for making claims and demonstrations about possible relationships and mechanisms (cf., [9, 10]) as well as showing the arguments and evidence for those claims.

To understand the relationships among Universal Entities in the physical world, we turned to natural science [4]. For social entities, we could turn to social science. This is reasonable since we accept the position that social entities are "real". Moreover, to the extent that social science makes causal predictions, we can use those predictions to confirm the validity of Entities. For physical phenomena, this type of confirmation of Entities is known as scientific warrant. Because there is more uncertainty about social science models, we may express our lower level of confidence for social entities by referring instead to consistency warrant.

In sociology, there are several grand theories, or major theoretical frameworks. Social ontology is a central aspect of each of these theories because they propose theoretical constructs and relationships among the constructs. Here, we focus on Parsons's AGIL [17] which asserts that there are four essential attributes a society must have to endure:

---

ontologies for BFO has focused on commitments and obligations [21]. Other specific proposals have focused on contracts, economics, and social aspects of medicine [14].

- **Adaptive:** This describes the need to adjust to the environment. Both shelter and farming would be considered as part of the Adaptive dimension. It covers many of the human needs identified by [16].
- **Goal Oriented:** This requires specification and accomplishment of social goals, and would include regulations, laws, and politics.
- **Integration:** This describes cohesion of the social group such as through family, religion, and language.
- **Latency:** A social group must renew its customs, knowledge, and values for the next generation through education.

Parsons's work is an application of Systems Theory to sociology (cf., [5]) and is often described as structure-functionalist[6]. Following our analysis of functionality, we propose that the Function of an Independent Continuant produces (or prevents) a State Change in a specific Independent Continuant-Process pair [5].[7] Thus, we might say:

- The Function of a ladle is to carry liquids.
- The Functions of Court music are to entertain and to impress guests. (Integration)
- The Function of certain types of physical structures (e.g. a house) is to shelter the inhabitants. (Adaptive)
- The Function of the education subsystem is to transmit knowledge. (Latency)

This description of Parsons's work just scratches the surface; for instance, he has an extensive discussion of the function of the family. It may be possible to develop a structured version of his entire framework. However, we should also note that among sociologists, there is disagreement about the value of the AGIL system.

Our emphasis on realism for social entities is also relevant to anthropology. We might first focus on the social science perspective to anthropology rather than the humanities perspective [18] (cf., [20]). Thus, we might emphasize archaeology and physical anthropology. Nonetheless, many entities and social activities such as rituals and icons that are the subject of anthropology clearly have deep symbolic, aesthetic, and emotional significance which we would need to account for.

## 6 Models of Particulars

One of the main goals of our direct representation approach is to provide highly-structured descriptions of specific cultures and histories. BFO defines Histories as a type of Occurrent [9]. Histories are said to be all the Processes associated with a given

---

[6] A full Functionalist model could have a web of Functions that address Needs. Mechanisms which satisfy Needs may themselves generate new Needs. BFO seems to lean toward a Structuralist view but its inclusion of Procedures with an object-oriented flavor suggests it could become more Functionalist.

[7] For an internally consistent ontology/model, all terms in the definitions should also be included in the ontology.

Continuant. This is an elegant definition but it tells us little about the relationships among those Processes.

Much of the discussion about the nature of historical explanations revolves around the notion from Hempel of a "covering law" which requires that any change should be justified (i.e., covered) by a law or reason for its occurrence. The expectation that there should be broad covering laws to account for events in history and culture has been widely criticized. Instead, Roberts [19] proposes that most major historical events (e.g., revolutions) do not have a single over-arching covering theory but are composed of smaller events each of which can be accounted for with covering theories. [12] makes a similar point, that claims about causal relationships among social phenomena need to be supported by models of mechanisms for how the entities interact.[8]

Because social situations are complex and because Thick Entities are generally composed of many parts it may be difficult to confirm causal processes. For instance, while it is easy to believe that the prosperity in the Roman Empire during the reign of the Antonine Emperors was due to their good policies [13], we cannot make that case with scientific rigor. After documenting the evidence, we may apply the generalization only while retaining some caution about it.

## 7   Repositories and Knowledgebases

Just as [4] proposed a variety of interrelated repositories for scientific research, similar interlocking repositories will be needed to complement the foundry of everyday activities and infrastructures. There would be several layers of knowledge resources:

- **Ontology and Model Foundry:** The everyday Human Activities and Infrastructures Foundry would include not only ontologies but also models of Thick Entities. The complete Foundry will require details of many different types of Procedures. In addition to the ontologies, the Foundry might include Reference Models such as of Bronze Age communities or Midwestern U.S. towns.

    We may not have full confidence in some of the Universals because there are competing theoretical frameworks. Thus, we may allow alternative representations using several of those frameworks. Related to this, we may apply a weaker consistency warrant rather than scientific warrant as a criterion for inclusion.

    Ontologies based on the BFO can be considered a type of classification system; after all, each BFO ontology is a taxonomy. A collection of BFO ontologies (i.e., a Foundry) can be viewed as an entity-based faceted classification.[9]

---

[8] Much of what is termed systems analysis appears focused more on process re-engineering than on systematic analysis of existing systems. Case studies can support what might properly be considered as systems analysis. Specifically, convergent case studies can be useful to evaluate possible causal mechanisms [12].

[9] The links of other entities (such as Locations, Dependent Continuants, and Processes) to the Object forms a sort of faceting. Indeed, it is easy to see the similarity to Raganathan's PMEST and to FrameNet's Frame Elements [2]. However, such entity-based faceting should be distinguished from other faceted classification systems which are subject based.

- **Models of Particulars:** See Section 6 above.

- **Primary Source Materials:** [3] called for cleaned and consistent repositories of historical source material. Moreover, these materials should have standard markup. In addition, databases of locations, climate, records, economic data, census reports, sports scores can also be coordinated with the Foundry ontologies and models.

- **Evidence, and Argumentation:** The evaluation of internal and external validity was a major factor in our discussion of scientific research reports [4]. We should have a similar focus here. For instance, [12] describes issues for the use of case studies in social science; we could develop an argumentation schema to organize and save evaluations of the validity of case studies.

- **Annotations, Secondary Materials, and Indexes:** Given that the foundry should be coordinated with other relevant repositories, we should allow annotations and include secondary materials. Potentially, structured direct representation would support many services such as supporting text and narrative generation for discourse functions such as explanation and argumentation.

## 8 Discussion

We have examined issues for collecting and coordinating applied ontologies and models for Everyday Human Activities and Infrastructures. These ontologies and models build on the rigorous semantics of the BFO and extend the constraints of BFO to everyday infrastructures and then to social and cultural descriptions. To do that, we relax some of the constraints but we expect that these will be flagged appropriately. This effort is as much about developing a useful information resource as about maintaining the purity of the ontological framework.

## References

1. Allen, R.B.: Toward an Interactive Directory for Norfolk, Nebraska: 1899-1900, IFLA Newspaper and Genealogy Section Meeting, Singapore, 2013. arXiv: 1308.5395.
2. Frame-based Models of Communities and their History. *Histoinformatics*, LNCS 8359, 2014, 110-119, doi: 10.1007/978-3-642-55285-4_9
3. Allen, R.B.: Issues for Direct Representation of History, ICADL, 2016, 218-224, doi: 10.1007/978-3-319-49304-6_26
4. Allen, R.B.: Rich Semantic Models and Knowledgebases for Highly-Structured Scientific Communication, 2017, arXiv: 1708.08423
5. Allen, R.B.: Rich Semantic Modeling, in preparation.
6. Allen, R.B., Song, H., Lee, B.E., Lee, J.Y.: Describing Scholarly Information Resources with a Unified Temporal Map, ICADL 2106, 212-217, doi: 10.1007/978-3-319-49304-6_25
7. Arp, R., Smith, B., Spear, A.D.: Building Ontologies with Basic Formal Ontology, MIT Press, Cambridge MA, 2015, also see http://purl.obolibrary.org/obo/bfo/Reference
8. Conway, S.: Thai Textiles. British Museum Press, London, 1992.
9. Chu, Y.M., Allen. R.B.: Formal Representation of Socio-Legal Roles and Functions for the Description of History, TPDL, 2016, 379-385, doi: 10.1007/978-3-319-43997-6_30

10. Diamond, J.: Guns, Germs, and Steel, Norton, New York, 1997.
11. Gasser, L.: Information and Collaboration from a Social/Organizational Perspective, S.Y. Nof (ed.), Information and Collaboration Models of Integration, 237-261, Kluwer, 1994.
12. George, A.L., Bennett, A.: Case Studies and Theory Development in the Social Sciences, MIT Press, Cambridge MA, 2004.
13. Gibbon, E.: The History of the Decline and Fall of the Roman Empire (1845). /www.gutenberg.org/files/731/731-h/731-h.htm
14. Jansen, L.: Four Rules for Classifying Social Entities, In: Philosophy, Computing and Information Science, 2014, London: Pickering & Chatto, pp 189-200
15. Lee, B.W., Lee Y.S.: (eds): Music of Korea, National Center for Korean Traditional Performing Arts, Seoul, 2007.
16. Maslow, A.H.: A Theory of Human Motivation. Psychological Review, 50, 370-396, 1943.
17. Parsons, T.: The Structure of Social Action, Free Press, Boston, 1968.
18. Peregrine, P., Moses, Y.T., Goodman, A., Lamphere, L., Peacock, J..L.: What Is Science in Anthropology? American Anthropologist, 114, 593–597, doi:10.1111/j.1548-1433.2012.01510.x
19. Roberts, C.: The Logic of Historical Explanation. Pennsylvania State University Press, State College PA, 1995.
20. Schilbrack. K.: A Realist Social Ontology of Religion, Journal of Religion, 27, 2017, 161-178, doi: 10.1080/0048721X.2016.1203834
21. Smith, B.: Searle and De Soto: The New Ontology of the Social World. In The Mystery of Capital and the Construction of Social Reality, B.Smith, D. Mark, I. Ehrlich (eds), Open Court, 2015, http://ontology.buffalo.edu/document_ontology/Searle&deSoto.pdf
22. Smith, B., Ashburner, M., Rosse, C., Bard, J., Bug, W., Ceusters, W., Goldberg, L.J., Eilbeck, K., Ireland, A., Mungall, C. J., Leontis, N., Rocca-Serra, P., Ruttenberg, A., Sansone, S.A., Scheuermann, R.H., Shah, N., Whetzel, P.L., Lewis, S.: The OBO Foundry: Coordinated Evolution of Ontologies to Support Biomedical Data Integration, Nature Biotechnology, 25, 1251–1255, 2007. doi: 10.1038/nbt1346